\documentclass[10pt,twocolumn,letterpaper]{article}

\usepackage[pagenumbers]{cvpr}

\usepackage{caption}
\usepackage{microtype}
\usepackage[export]{adjustbox}
\usepackage{array,tabularx}
\usepackage{needspace}
\usepackage{placeins}
\definecolor{projectblue}{RGB}{0,114,189}

\makeatletter
\renewcommand\paragraph{\@startsection{paragraph}{4}{\z@}%
  {5pt plus 1pt minus 1pt}{-1em}{\normalfont\normalsize\bfseries}}
\makeatother

\definecolor{cvprblue}{rgb}{0.21,0.49,0.74}
\usepackage[pagebackref,breaklinks,colorlinks,allcolors=cvprblue]{hyperref}

\def\confName{CVPR}
\def\confYear{2027}

\title{BioAgent-World: Interactive Digital Twins for Scalable Robot Skill Learning}

\author{Gaojing Zhang$^{1,2}$,
Changwen Chen$^{1}$,
Zichen Luo$^{1}$,
Yuhan Wang$^{1}$,
Jialiang Li$^{1}$,
Wenzhao Lian$^{1,{\dagger}}$
\\[3pt]
{\small
$^{1}$Shanghai Jiao Tong University
\qquad
$^{2}$University of Sussex
}
\\[2pt]
{\footnotesize
$^{\dagger}$Corresponding author
}}

\begin{document}
\raggedbottom
\twocolumn[{%
\renewcommand\twocolumn[1][]{#1}%
\maketitle
\vspace{-15pt}

\begin{minipage}{\textwidth}
    \captionsetup{hypcap=false}
    \centering
    \adjustimage{trim={0.003\width} {0.004\height} {0.003\width} {0.004\height},clip,width=0.95\textwidth}{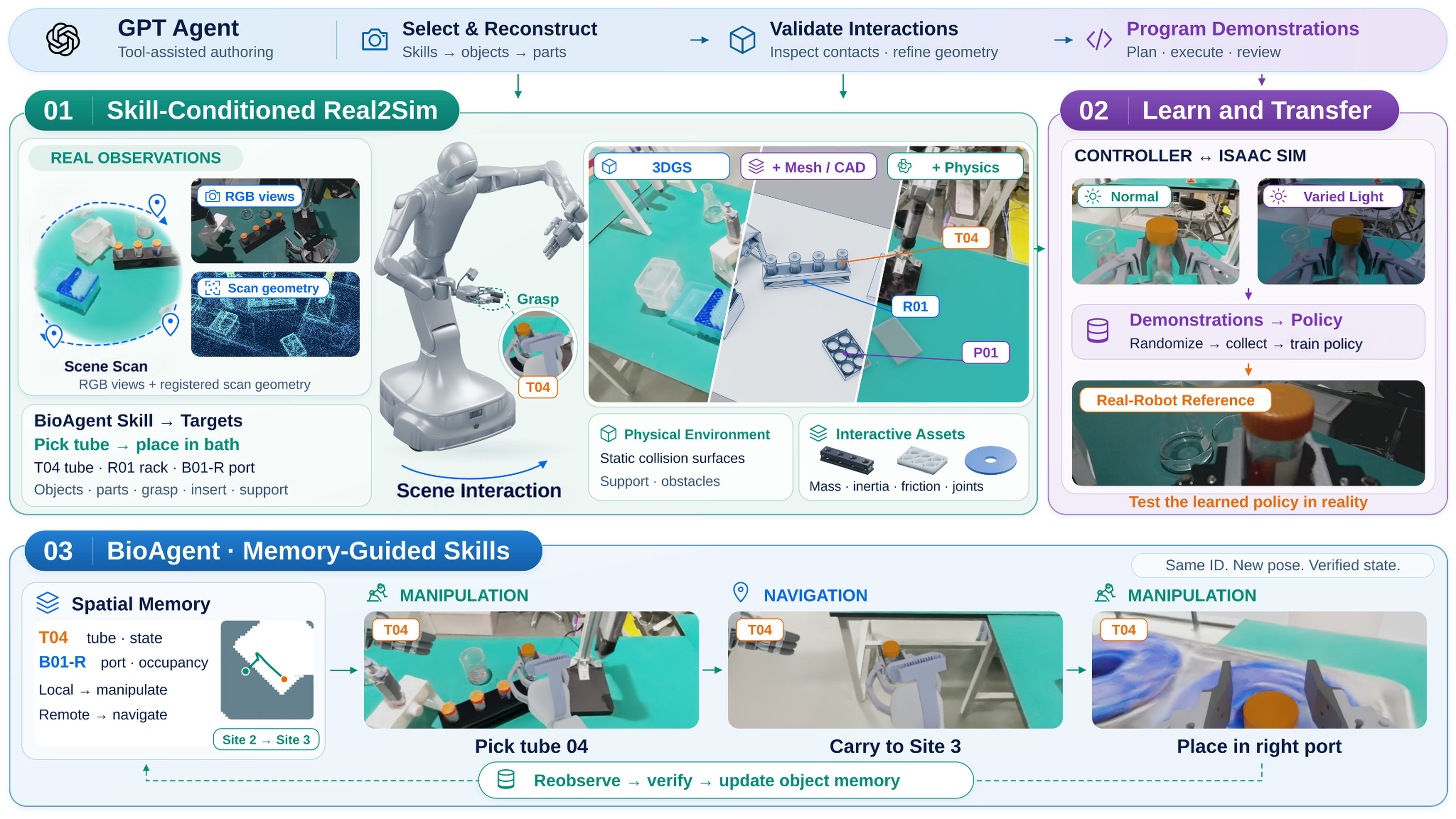}
    \captionof{figure}{
        \textbf{Overview of BioAgent-World.}
        Skill requirements guide the reconstruction and interaction
        validation of real scenes. The resulting digital twins support
        demonstration generation and policy learning, while persistent
        object and state memory supports manipulation and navigation
        across workstations.
    }
    \label{fig:teaser}
\end{minipage}%
\par\vspace{10pt}
}]

\begin{abstract}
Collecting high-quality demonstrations on real robots is costly, motivating the use of simulation to scale up robot skill learning.
However, visually accurate reconstructions can still misrepresent the physical interactions and state transitions required for real-world execution.
To bridge this gap, we present BioAgent-World, a framework that constructs interactive digital twins from real observations for scalable robot skill learning.
Rather than relying solely on visual appearance, our framework infers geometric, mechanical, and functional scene properties and verifies them through physical measurements and interaction tests. When observations are ambiguous, it actively refines the scene by selecting new viewpoints that most efficiently reduce task-execution uncertainty.
Guided by these validated twins, a skill agent generates demonstrations under constrained randomization, while persistent object-state memory ensures multi-step consistency and feeds real-world execution discrepancies back into model refinement.
Across 11 laboratory skills, three policy backbones, and 1,320 real-world skill trials, policies trained only on our synthetic data retain 87.39--92.22\% of the macro-averaged success rate of real-data policies under matched training-data budgets.
Furthermore, BioAgent-World uses eight NVIDIA RTX 4090 GPUs for rendering and data collection, generating 2,200 accepted trajectories in 7.54 hours, compared to 36.8 hours for real collection. These results establish interaction-validated digital twins as a practical, scalable data engine for real-world robot manipulation in laboratory environments.  Project page: {\normalfont 
{\color{projectblue}\urlstyle{same}\url{https://suiuko.github.io/BioAgent_worldweb/}}.}

\end{abstract}
\section{Introduction}
\label{sec:introduction}

Visuomotor policies learn manipulation behaviors from observation--action data~\cite{baw_chi2023diffusionpolicy,baw_pi2025pi05}, but collecting demonstrations on real robots remains costly. Laboratory workflows severely compound this burden: they involve contact-rich, multi-stage procedures that require frequent manual resets whenever a failure occurs. For example, retrieving a tube, opening its cap, and aspirating liquid each depend on the exact contact geometry, physical constraints on object motion, and functional states caused by preceding actions. Simulation offers a promising way to scale training data without repeated manual operation. MimicGen~\cite{baw_mandlekar2023mimicgen} adapts a small set of human demonstrations to new task instances, whereas RoboTwin~2.0~\cite{baw_chen2026robotwin2} generates expert programs with simulation feedback and applies domain randomization. However, the real-world utility of synthetic data hinges on whether simulated physical interactions faithfully transfer to real robots.

To ground simulation in reality, recent real-to-sim methods reconstruct physical environments directly for policy learning. RialTo~\cite{baw_torne2024rialto} refines policies in digital twins of real scenes, while automated digital-cousin generation~\cite{baw_dai2025digitalcousins} creates related scenes with similar geometric and semantic affordances. For precise laboratory manipulation, visual fidelity alone is insufficient: reconstructed objects must also preserve the geometry and functional states required for interaction. For example, a tube rack may appear correct while its collision geometry prevents tube insertion. Similarly, a tube may appear uncapped while the simulator still treats it as closed, preventing subsequent aspiration. Such hidden discrepancies may alter physical interactions, including contact behavior and relative motion between connected parts, as well as state transitions. Closing this gap requires linking reconstruction to measurable skill conditions, preserving functional-state consistency across actions, and identifying unresolved scene properties that warrant further real-world observation.

To address these challenges, we present BioAgent-World, a framework in which task requirements guide the construction and validation of interactive digital twins for scalable robot skill learning. Specifically, BioAgent-World makes three contributions:
\begin{itemize}
    \item We construct interactive laboratory scenes and simulation assets with visual and collision geometry, physical properties, and joints that define relative motion between parts. The assets also track functional states, including cap status, container contents, and liquid volume, so that successive skills can account for changes caused by earlier actions.

    \item We develop a pipeline for scene generation, refinement, and validation guided by skill requirements. When candidate scene models disagree on whether a skill can be executed, the pipeline selects additional real-world observations by their expected reduction in decision error per unit cost. The resulting evidence guides local corrections, followed by interaction tests to validate the refined assets.

    \item We adapt an agent system for long-horizon laboratory protocols to the reconstructed simulation environment for task planning and coordinated skill execution. Experiments demonstrate that the environment supports the execution of interdependent laboratory skills and that the skill sequences orchestrated in simulation can be transferred directly to the real robot.
\end{itemize}

\section{Related work}
\label{sec:related_work}

\subsection{Interactive real-to-sim reconstruction}
\label{sec:related_work:real2sim}

Recent real-to-sim pipelines advance robot learning along two primary directions: scene-level digital twinning and the generation of simulation assets with parts connected by joints. At the scene level, methods like RialTo~\cite{baw_torne2024rialto} and Digital Cousins~\cite{baw_dai2025digitalcousins} reconstruct real environments or their geometric variants to enable policy refinement in simulation. At the asset level, recent frameworks incorporate visual-geometric checks~\cite{baw_huang2026literealityagent}, physical simulation feedback~\cite{baw_cai2026artisan}, and executable behavior contracts~\cite{baw_song2026chemlabassets} to model objects with movable parts and laboratory instruments.

However, adapting these reconstructions to laboratory workflows requires validating skill-specific contacts and state transitions—a step often omitted in prior work, which either lacks ground-truth physical parameter validation~\cite{baw_huang2026literealityagent} or stops short of real-hardware contact testing~\cite{baw_song2026chemlabassets}. To bridge this gap, BioAgent-World explicitly links skill preconditions to underlying scene properties and real-world observations. When simulated executions diverge from physical evidence, these discrepancies trigger targeted measurements and local refinements, yielding formal validation records that bound downstream data generation.

\subsection{Robot training data generation}
\label{sec:related_work:data_generation}

Scaling robot training datasets typically relies on either replaying segmented human demonstrations in novel contexts~\cite{baw_mandlekar2023mimicgen, baw_jiang2025dexmimicgen} or synthesizing tasks via domain randomization and simulation feedback~\cite{baw_chen2026robotwin2, baw_nasiriany2026robocasa365}.

Although these approaches are effective for general manipulation, generating valid data in reconstructed laboratories requires faithful representations of the contact interfaces and functional states involved in real-world tasks. BioAgent-World enforces this fidelity throughout data collection. Before execution, it restricts sampling to validated regions through precondition checks and admissible randomization. During and after rollout, it maintains functional-state consistency across skills and filters out trajectories that contain physical artifacts or observation--action timing mismatches.

\subsection{Task-driven simulation refinement}
\label{sec:related_work:refinement}

Active exploration and simulation adaptation use real-world feedback to improve sim-to-real transfer. Liang et al.~\cite{baw_liang2020active} learn exploration policies that identify task-relevant aspects of how the physical system responds to actions. These policies are optimized for downstream task performance. ASID~\cite{baw_memmel2024asid} uses Fisher information to guide exploration and refines simulator properties from real interaction data. AdaptSim~\cite{baw_ren2023adaptsim} updates simulation parameter distributions from task-policy feedback to improve performance in the target environment.

Rather than identifying global parameters or optimizing end-to-end policies, BioAgent-World focuses on resolving uncertainty over whether a reconstructed scene satisfies specific skill preconditions. Using a fixed diagnostic controller, the system first detects where observation-compatible scene hypotheses disagree on condition satisfaction. It then actively selects real-world measurements that maximally reduce the risk of misjudging skill usability per unit cost. Finally, this newly acquired evidence updates the scene hypotheses, guides local parameter corrections, and re-establishes valid bounds for data generation.

\section{Method}
\label{sec:method}

BioAgent-World constructs skill-conditioned interactive digital twins by converting observed objects into editable assets with contact geometry, joints that define relative motion between parts, physical properties, and functional states. Skill requirements determine which properties must be represented and checked. Discrepancies between simulated execution and real-world evidence guide targeted observations and local refinement. Validated assets support demonstration generation, while persistent object and state memory supports execution across workstations and feeds real-world evidence back into reconstruction.

\begin{figure*}[!t]
    \centering
    \adjustimage{trim={0.006\width} {0.005\height} {0.006\width} {0.005\height},clip,width=\textwidth}{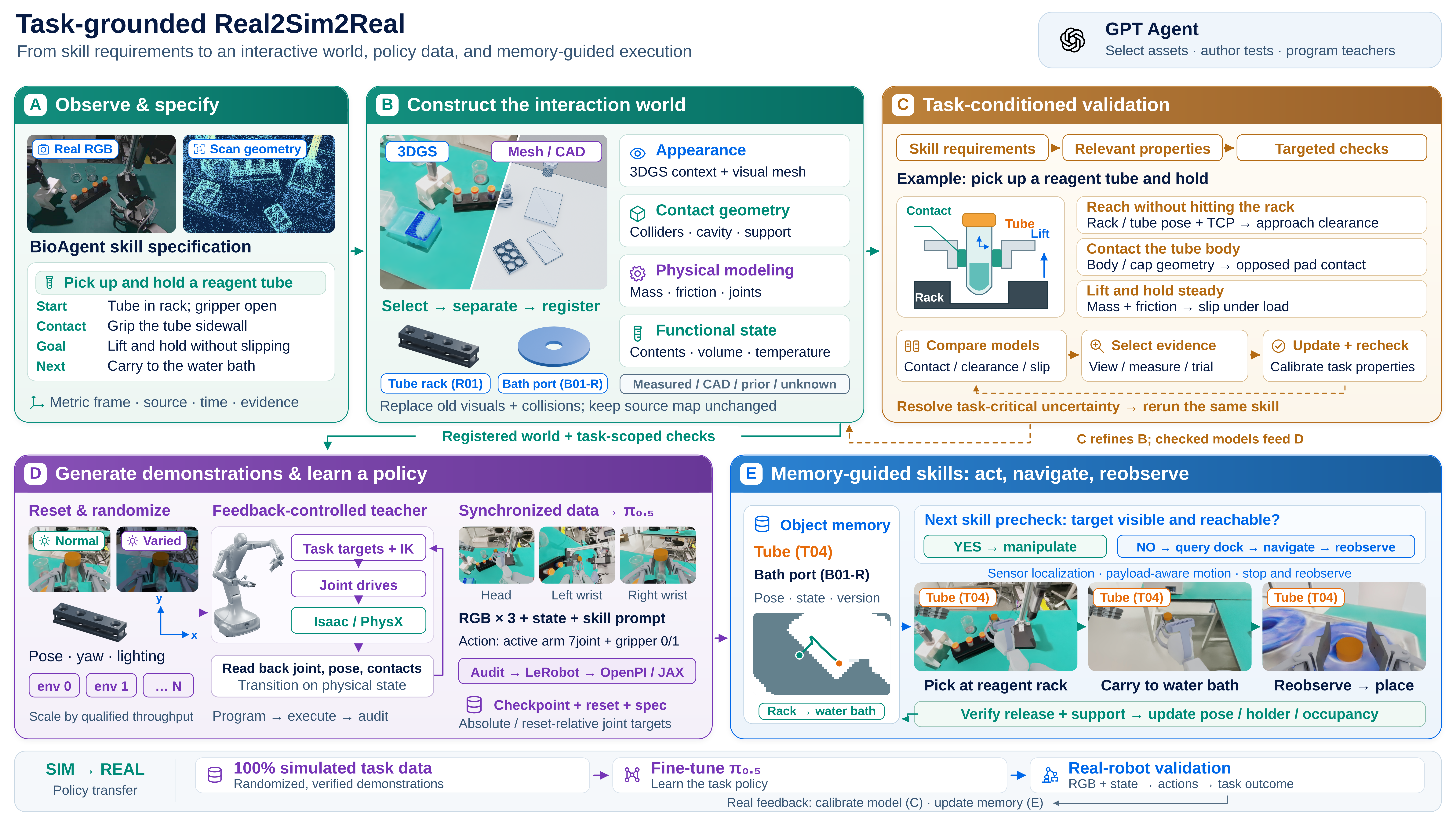}
    \caption{
        \textbf{Workflow of BioAgent-World.}
        (A) Real observations and skill specifications define the
        reconstruction requirements.
        (B) Scene construction models appearance, contact geometry,
        how objects respond to robot actions, joints that constrain
        relative motion between parts, and functional states.
        (C) Task-conditioned validation guides evidence acquisition
        and local refinement.
        (D) Validated scenes support demonstration generation
        and policy learning.
        (E) Persistent object and state memory supports manipulation,
        navigation, and reobservation, with execution feedback
        informing model refinement and memory updates.
    }
    \label{fig:method_overview}
\end{figure*}

Figures~\ref{fig:teaser} and \ref{fig:method_overview} show
the system overview and detailed workflow, respectively.

\subsection{Problem Formulation and World Representation}
\label{sec:method:representation}

We construct an interactive world from environment observations $\mathcal{D}$, the robot's URDF and calibration $\mathcal{R}$, and skill specifications $\mathcal{K}$. The resulting world supports laboratory workflow simulation and training data generation for supported skills, while also providing a foundation for task execution in the partially observed real environment.
We represent the world as
\begin{equation}
\mathcal{W} = \bigl(M_{\mathrm{obs}}, M_{\mathrm{vis}}, C_{\mathrm{col}},
\mathcal{O}, \mathcal{S}, \mathcal{E}\bigr),
\label{eq:baw:world}
\end{equation}
where $M_{\mathrm{obs}}$ preserves observed geometry. The visual layer $M_{\mathrm{vis}}$ stores renderable appearance and identified completions, while the collision layer $C_{\mathrm{col}}$ stores derived collision geometry. Each object in $\mathcal{O}$ has a persistent identity and a pose with uncertainty. Its record includes geometry, joints connecting its parts, and physical parameters, together with functional state, affordances, and evidence. The skill registry $\mathcal{S}$ binds each specification to its target identities, preconditions, and postconditions. Each entry also specifies an executor, a docking pose, an execution profile defining skill-specific operating constraints, and a data/model interface contract. Event records $\mathcal{E}$ store observations and executions with timestamps, domains, and versions. They distinguish among real measurements, native simulator readbacks, synthetic test fixtures, and model-maintained estimates of contents or volume.

\subsection{Skill-conditioned scene construction}
\label{sec:method:construction}

\paragraph{Room and interaction scales.}
The room-level representation captures macro-scale structural elements, including walls, work surfaces, large equipment, passages, and localization constraints. Derived from room scans, this spatial layout bounds the placement of task-relevant objects and establishes the geometric context required for workstation-level navigation and manipulation.

\paragraph{Interactive asset reconstruction.}
Within the reconstructed room, we instantiate each task-relevant object as an independently editable asset linked to its persistent identity in $\mathcal{O}$. While raw observations initialize object geometry and poses, documented CAD models and physical measurements provide fine-grained details required by target skills. To support both rendering and physical interaction, each asset maintains decoupled visual and collision representations; notably, the collision mesh preserves critical contact surfaces and free spaces, such as the cavities and support surfaces of a tube rack.

Beyond static geometry, we configure each asset's physical dynamics and semantic state transitions. Specifically, we assign inertial and frictional parameters to rigid bodies and define kinematic joints to constrain motion between articulated parts. Non-geometric properties, such as cap status, container contents, and liquid volume, are tracked explicitly as object states. During simulation, the physics engine resolves contacts and joint motion, while the accompanying object record updates these functional states. Finally, binding these physical and semantic components to registered skill targets enables automated verification of whether an asset satisfies its required interaction constraints.

\paragraph{Compiling skill conditions.}
To make the constructed assets testable against skill requirements, we map preconditions and postconditions to measurable properties through a typed predicate library, including \texttt{inside}, \texttt{supported}, \texttt{grasp\_accessible}, and \texttt{unlocked\_before\_open}. Each predicate specifies admissible participant types, required properties and observations, and verification tools that evaluate satisfaction or identify missing evidence.

We trace these conditions across object parts and spatial relations to identify the specific model properties and empirical evidence required for evaluation. In tube placement, for example, these properties encompass cavity topology, relative pose, support height, and grasping clearance. We conservatively retain all resulting dependencies, prioritizing those that trigger disagreements on skill feasibility across candidate models.

\paragraph{Geometric and temporal conditions.}
For a candidate model $M$ of the scene and robot calibration, evaluated over an execution trace $\tau$, each evaluable continuous condition $j$ has a signed margin $g_j(M,\tau)$: positive values indicate satisfaction, negative values indicate violation, and zero marks the constraint boundary. For instance, tube placement requires sufficient clearance for insertion and enough exposed tube length for later retrieval. Assuming a cylindrical rack opening and a tube with parallel axes, illustrative margins are
\begin{equation}
\begin{aligned}
g_{\mathrm{radial}} &= r_{\mathrm{hole}}-r_{\mathrm{tube}}-e_{\mathrm{lateral}},\\
g_{\mathrm{regrasp}} &= \ell_{\mathrm{exposed}}-\ell_{\mathrm{required}},
\end{aligned}
\label{eq:baw:margins}
\end{equation}
where $r_{\mathrm{hole}}$ and $r_{\mathrm{tube}}$ are the opening and tube radii, $e_{\mathrm{lateral}}$ is the lateral offset between their axes at insertion, $\ell_{\mathrm{exposed}}$ is the tube length above the rack after placement and $\ell_{\mathrm{required}}$ is the minimum exposed length for gripper-pad access. Validation additionally considers cavity structure, object tilt, swept-path collisions, and neighboring objects; discrete requirements, such as cavity existence, are checked separately.

For execution-dependent conditions, temporal monitors following specification-based robot validation~\cite{baw_li2026validating} check action ordering and persistent states, such as support remaining after gripper release. Simulated checks use simulator states, while real-world checks use available images, robot signals, or device feedback. Conditions without sufficient observations remain unresolved and motivate further measurements in Sec.~\ref{sec:method:refinement}.

\begin{figure*}[!t]
    \centering
    \includegraphics[width=0.96\textwidth]{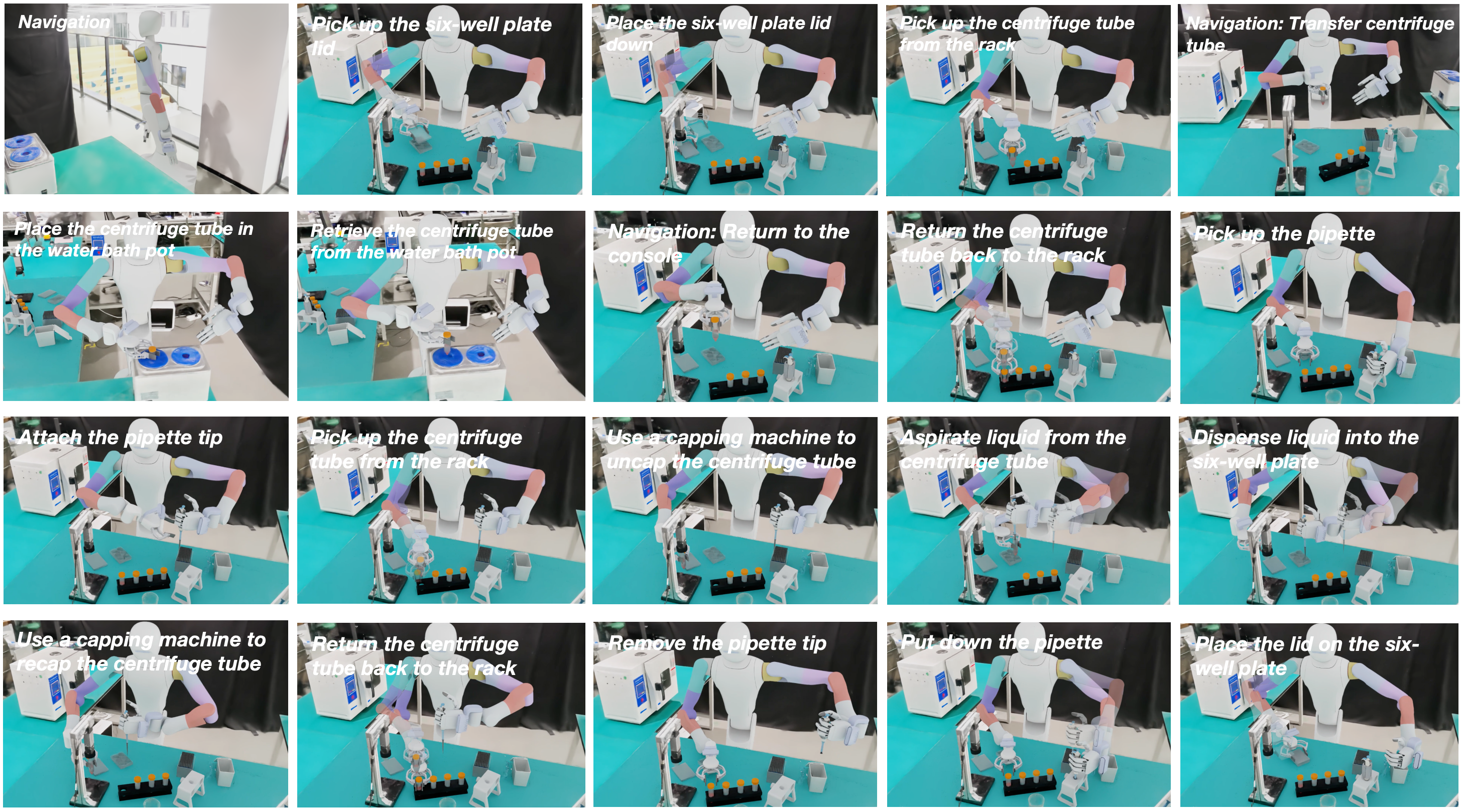}
    \caption{
    \textbf{Simulated execution of laboratory skills in BioAgent-World.}
    Representative snapshots show navigation between workstations,
    reagent-tube transport and cap manipulation, six-well plate
    handling, and liquid aspiration and dispensing with a micropipette
    within the reconstructed digital twins.
    }
    \label{fig:sim_robot_tasks}
\end{figure*}

\subsection{Interaction-guided refinement and validation}
\label{sec:method:refinement}

Observation-compatible reconstructions may predict different outcomes for the same skill. We use these disagreements to select targeted real-world evidence and refine the implicated scene or calibration parameters.

\paragraph{Observation-compatible hypotheses.}
We maintain a finite, weighted set of candidate models approximating
\begin{equation}
p\bigl(z,\theta_{\mathrm{scene}},\eta_{\mathrm{robot}}\mid\mathcal{D}\bigr),
\label{eq:baw:hypotheses}
\end{equation}
where each candidate $M=(z,\theta_{\mathrm{scene}},\eta_{\mathrm{robot}})$ comprises a structural alternative $z$, scene parameters $\theta_{\mathrm{scene}}$, and uncertain robot calibration parameters $\eta_{\mathrm{robot}}$. The scene parameters specify registration, how joints constrain relative motion between connected parts, and physical properties. Each model is consistent with reliable observations and retains the provenance of its measurements, CAD geometry, and prior assumptions.

For each skill test $s$, comprising a specified skill, initial state, and perturbation, we evaluate all candidates with the same diagnostic controller and observation interface. Following counterexample-guided refinement~\cite{baw_clarke2000cegar}, we compare condition margins, contract outcomes, and subsequent-skill feasibility to locate task-relevant disagreements.

\paragraph{Skill-usability decision risk.}
Let $Y_s(M)\in\{0,1\}$ indicate whether skill test $s$ satisfies its contract under model $M$. We quantify the risk of accepting or rejecting a skill incorrectly as
\begin{equation}
R(p)=\sum_s w_s\min_{d\in\{0,1\}}
\mathbb{E}_{M\sim p}\!\left[\ell\bigl(d,Y_s(M)\bigr)\right],
\label{eq:baw:risk}
\end{equation}
where $p(M)$ is the current candidate distribution, $d$ is a binary usability decision, $w_s$ weights task relevance, and $\ell$ penalizes incorrect decisions. Under unit symmetric binary loss, the per-test risk is $\min\{P(Y_s=1),1-P(Y_s=1)\}$, reflecting disagreement about skill usability. 

\paragraph{Selecting informative observations.}
Following expected-error-reduction methods~\cite{baw_roy2001errorreduction}, we score each admissible observation $q$ by its expected reduction in decision risk per unit cost:
\begin{equation}
V(q)=\frac{R(p)-\mathbb{E}_{y\sim p(y\mid q)}
\left[R(p_{q,y})\right]}{c(q)},
\label{eq:baw:query}
\end{equation}
where $y$ is a possible observation result, $p(y\mid q)$ its predicted distribution, $p_{q,y}$ the resulting posterior over models, and $c(q)>0$ the acquisition, reset, and tool cost. Queries include additional views, local depth, dimension measurements, state readbacks, and interaction trials.

\begin{figure*}[!t]
    \centering
    \includegraphics[width=0.96\textwidth]{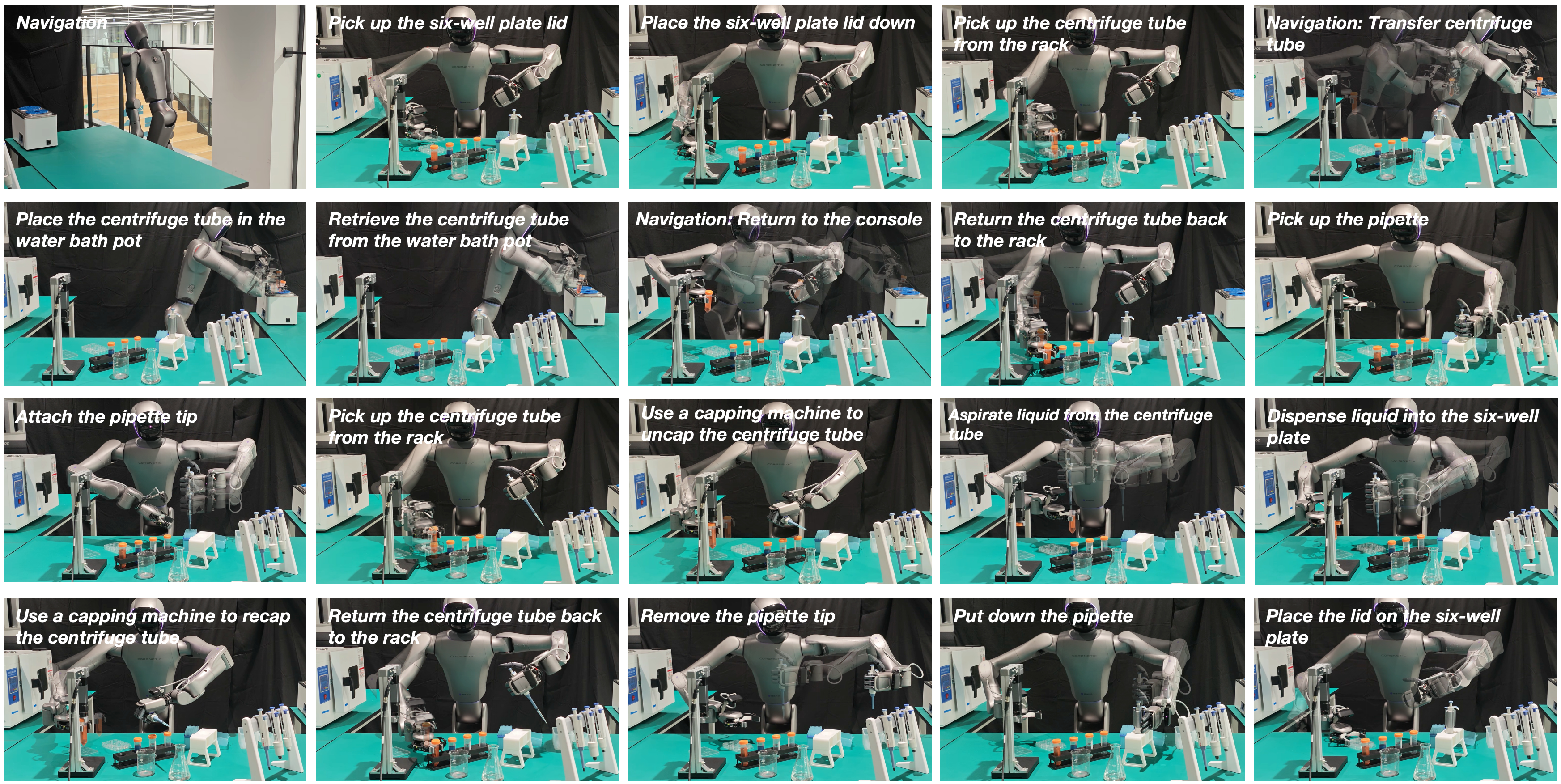}
    \caption{
    \textbf{Real-world execution of laboratory skills.}
    Representative snapshots show the robot navigating between
    workstations, handling six-well plates, transporting reagent
    tubes, opening and closing tube caps with a capping device,
    and aspirating and dispensing liquid with a micropipette.
    }
    \label{fig:real_robot_tasks}
\end{figure*}

We choose the highest-value admissible query within budget, hardware, and authorized-trial limits, targeting properties implicated by the compiled skill conditions~\cite{baw_liang2020active,baw_memmel2024asid,baw_ren2023adaptsim}. Geometric screening precedes short native-physics rollouts; when full risk evaluation is costly, a candidate-pair approximation prioritizes queries that distinguish models with conflicting skill outcomes.

\paragraph{Updating and scoped validation.}
After obtaining real evidence $y_{\mathrm{real}}$ from query $q$, we update model weights by
\begin{equation}
p_{\mathrm{new}}(M)\propto
p(y_{\mathrm{real}}\mid q,M)\,p_{\mathrm{old}}(M),
\label{eq:baw:update}
\end{equation}
where the likelihood reflects measurement uncertainty. Evidence also guides local corrections using robust residual fitting for visual registration and search or reweighting for physical parameters.

We revalidate the assets through same-view and held-out-view comparisons, measurements, and contact trials, checking coordinate consistency, independent parts, valid joints, and compiled skill conditions. Validation records delimit the tested skills and operating conditions under which the reconstructed scene supports demonstration generation (Sec.~\ref{sec:method:generation}).

\subsection{Task Planning and Data Collection}
\label{sec:method:generation}

We adapt BioAgent as our task-planning and data-collection framework within the reconstructed simulation environment, building directly on the interactive assets and validated skill preconditions from Sec.~\ref{sec:method:refinement}.

\paragraph{Task planning and supervised execution.}
BioAgent uses three subagents for task planning and execution in simulation. The Planner receives a task instruction and decomposes it into a sequence of registered skills. The Executor follows this plan to execute the skills and collect data in the simulator, while the Monitor supervises the Executor. Each skill is bound to its target objects; preconditions are checked before execution and postconditions afterward. Within a multi-skill episode, object and functional states carry over between skills and are used to check subsequent preconditions.

\paragraph{Demonstration collection.}
We use a server equipped with eight NVIDIA RTX 4090 GPUs for simulation rendering and demonstration collection.
Collection begins with a valid scene reset within the applicable validation record's scope. Each independent reset starts a new episode and invalidates the previous episode's plan tokens and state information that can change during execution. Randomization separates residual calibrated uncertainty from intended layout changes, control or localization perturbations, and boundary recovery conditions. During execution in the physical simulator, we record three camera streams, robot commands, and events with their clocks and interface versions. Trajectories are accepted only after passing checks for contract satisfaction, physical validity, observation--action timing, and consistency among visual, collision, and functional states. 

\subsection{Memory-grounded real-world transfer}
\label{sec:method:transfer}

\paragraph{Persistent object and state memory.}
While prior scene graphs~\cite{baw_hughes2022hydra, baw_gu2024conceptgraphs} and spatial memory systems~\cite{baw_huh2026homebody} focus primarily on geometric perception and skill reuse, our persistent memory additionally tracks dynamic task states, storing object identities, functional states, affordances, observation timestamps, and containment or support relations.

\paragraph{Docking and transport.}

For workstation manipulation, we validate docking poses against kinematic reachability, joint limits, camera visibility, whole-body and payload collision clearance, and exit-path availability. To govern object transport, we define a \emph{carry profile} encoding valid grasps, object geometry and mass, orientation bounds, velocity and acceleration limits, and the whole-body footprint, while respecting base rolling constraints (including nonholonomic fixed-wheel kinematics). Across the entire transport motion, we verify the carrying posture and 3D swept volume from withdrawal to final placement, enforcing strict tilt and angular-velocity limits on open containers.

\begin{table*}[!t]
\centering
\caption{Real-world skill execution by policies trained on equal amounts of real data (Real) or BioAgent-World data (BAW), compared within each backbone. Each skill is initialized independently before every trial. Skill entries report successes/20, while aggregate entries report successes/220 (macro-averaged SR in \%). The SR gap is Real minus BAW in percentage points, and performance retention is BAW divided by Real, computed from unrounded macro-averaged SRs.}
\label{tab:experiments:skill_transfer}
\small
\setlength{\tabcolsep}{3pt}
\begin{tabularx}{\textwidth}{@{}>{\raggedright\arraybackslash}X*{6}{c}@{}}
\toprule
& \multicolumn{2}{c}{$\pi_{0.5}$}
& \multicolumn{2}{c}{Diffusion Policy}
& \multicolumn{2}{c}{LingBot-VA} \\
\cmidrule(lr){2-3}\cmidrule(lr){4-5}\cmidrule(lr){6-7}
Skill & Real & BAW & Real & BAW & Real & BAW \\
\midrule
1. Pick up the six-well plate lid
& 19/20 & 18/20
& 14/20 & 12/20
& 17/20 & 17/20 \\

2. Pick up the centrifuge tube from the rack
& 18/20 & 16/20
& 13/20 & 11/20
& 17/20 & 15/20 \\

3. Place the centrifuge tube in the water bath
& 15/20 & 14/20
& 10/20 & 8/20
& 14/20 & 14/20 \\

4. Retrieve the centrifuge tube from the water bath
& 18/20 & 17/20
& 10/20 & 9/20
& 16/20 & 13/20 \\

5. Return the centrifuge tube to the rack
& 16/20 & 16/20
& 9/20 & 8/20
& 17/20 & 15/20 \\

6. Uncap the centrifuge tube
& 15/20 & 14/20
& 9/20 & 7/20
& 14/20 & 12/20 \\

7. Aspirate liquid from the centrifuge tube
& 20/20 & 20/20
& 16/20 & 16/20
& 20/20 & 19/20 \\

8. Dispense liquid into the six-well plate
& 19/20 & 16/20
& 9/20 & 8/20
& 17/20 & 16/20 \\

9. Recap the centrifuge tube
& 12/20 & 11/20
& 6/20 & 4/20
& 13/20 & 10/20 \\

10. Grasp the plate lid from the table
& 19/20 & 17/20
& 12/20 & 11/20
& 17/20 & 16/20 \\

11. Place the lid on the six-well plate
& 9/20 & 7/20
& 3/20 & 3/20
& 10/20 & 9/20 \\
\midrule
Aggregate success
& \shortstack{180/220\\(81.82)}
& \shortstack{166/220\\(75.45)}
& \shortstack{111/220\\(50.45)}
& \shortstack{97/220\\(44.09)}
& \shortstack{172/220\\(78.18)}
& \shortstack{156/220\\(70.91)} \\

SR gap (percentage points)
& \multicolumn{2}{c}{6.36}
& \multicolumn{2}{c}{6.36}
& \multicolumn{2}{c}{7.27} \\

Performance retention (\%)
& \multicolumn{2}{c}{92.22}
& \multicolumn{2}{c}{87.39}
& \multicolumn{2}{c}{90.70} \\
\bottomrule
\end{tabularx}
\end{table*}

\paragraph{Execution and feedback.}
For each planned task step, we use the object and state memory to check object identity, functional state, observation freshness, and skill applicability. After navigation, the robot stops and acquires a fresh close-range observation. Using this observation, we recheck manipulation eligibility before invoking the skill agent. Fresh post-action evidence determines whether to advance to the next step or initiate recovery.

We register new observations of moved objects and associate them with object identities before updating state versions and invalidating dependent plans. With execution paused or controlled, we revise the derived visual and collision scene and verify it by readback before replanning (Sec.~\ref{sec:method:refinement}).

\section{Experiments}
\label{sec:experiments}

We evaluate BioAgent-World through three research questions. RQ1 asks whether data collected in the reconstructed environments support real-world skill execution, measured by downstream policy performance. RQ2 examines data-collection efficiency in terms of trajectory acceptance, total production time, throughput, and human involvement. RQ3 examines how the reconstructed environments support BioAgent in planning and executing long-horizon laboratory protocols. We additionally evaluate the contribution of major components through ablation studies, reported in the supplementary material (Appendix A).

\subsection{Experimental setup}
\label{sec:experiments:setup}

We compare two training-data sources across three policy backbones: $\pi_{0.5}$~\cite{baw_pi2025pi05}, Diffusion Policy~\cite{baw_chi2023diffusionpolicy}, and LingBot-VA~\cite{baw_li2026lingbotva}. For each backbone, we train policies separately on equal amounts of real data or BioAgent-World data while holding the architecture and robot task conditions fixed. We evaluate the resulting policies on the same 11 laboratory skills, covering six-well plate handling, tube placement and retrieval at the rack and water bath, tube-cap manipulation, and liquid transfer. Each skill is evaluated independently, with its initial state restored before every trial. We conduct 20 real-world trials per skill for each backbone--data-source condition, yielding 220 skill trials per condition and 1,320 across all six conditions.
Figure~\ref{fig:real_robot_tasks} shows representative real-world executions involving navigation, container handling, and pipetting. We run the reconstructed environments in NVIDIA Isaac Sim~\cite{baw_nvidia_isaacsim} and train our policies on eight NVIDIA A100 GPUs.
We initialize \(\pi_{0.5}\) from official pretrained weights, freezing the vision-language backbone and fine-tuning the action expert and projection layers. Training uses AdamW, a global batch size of 128, 20,000 steps, and a 32-step action horizon.
Diffusion Policy is trained from scratch per task using Adam, batch size 64, and 50,000 steps, with a learning rate of \(10^{-4}\).
LingBot-VA fine-tunes a pretrained video-action Transformer with frozen VAE and text encoder, using AdamW, batch size 64, 5,000 steps, and a learning rate of \(10^{-5}\) with FSDP.

We report per-skill success rate (SR) and the macro-averaged SR over the 11 skills. With equal trial counts across skills, this macro-average equals the pooled success count divided by 220. Two additional descriptive metrics summarize transfer: the SR gap, defined as real-data SR minus BioAgent-World SR in percentage points, and performance retention, defined as the BioAgent-World macro-averaged SR divided by the real-data macro-averaged SR. RQ2 evaluates simulation data production on a server equipped with eight NVIDIA RTX 4090 GPUs, with throughput computed as the number of accepted trajectories divided by total production time.
\subsection{RQ1: Can Generated Data Support Real-World Skills?}
\label{sec:experiments:rq1}

Table~\ref{tab:experiments:skill_transfer} shows that BioAgent-World data support real-world skill execution across all three tested backbones. 
The generated-data policies achieve macro-averaged SRs of 75.45\%, 44.09\%, and 70.91\% for $\pi_{0.5}$, Diffusion Policy, and LingBot-VA, respectively. The corresponding gaps relative to real-data policies are 6.36, 6.36, and 7.27 percentage points, with performance retention ranging from 87.39\% to 92.22\%. 

Transfer gaps vary by skill and backbone. For tube-cap opening and closing, generated-data SRs are 5--15 percentage points below those of their real-data counterparts. The largest observed per-skill gap is 15 percentage points, occurring for $\pi_{0.5}$ on liquid dispensing and for LingBot-VA on tube retrieval from the water bath and tube recapping. Final plate-lid replacement remains difficult under both data sources: generated-data policies achieve 15--45\% SR, compared with 15--50\% for real-data policies. These findings motivate targeted refinement of tube interactions, liquid dispensing, and lid placement. 

\subsection{RQ2: Does Simulation Reduce Collection Cost?}
\label{sec:experiments:rq2}

\paragraph{Matched collection protocol.}
We compare BioAgent-World with real teleoperation across the 11 RQ1 skills under identical task conditions. Each source provides 2,200 accepted trajectories (200 per skill). Total production time covers the entire process from preparation to dataset delivery, including setup, generation or recording, failed attempts, resets, screening, and export. BioAgent-World generates trajectories autonomously after an operator resolves errors that prevent task completion.

\begin{table}[!htbp]
\centering
\caption{Data-production comparison across 11 skills, with 200 accepted trajectories per skill and source. BioAgent-World uses eight NVIDIA RTX 4090 GPUs. Production time spans preparation through dataset delivery. }
\label{tab:experiments:production_throughput}
\small
\setlength{\tabcolsep}{3pt}
\begin{tabularx}{\columnwidth}{@{}>{\raggedright\arraybackslash}Xrr@{}}
\toprule
Measure
& \shortstack{BioAgent-\\World}
& \shortstack{Real\\Teleoperation} \\
\midrule
Candidates screened & 2,466 & 2,267 \\
Accepted trajectories & 2,200 & 2,200 \\
Rejected trajectories & 266 & 67 \\
Acceptance rate (\%) & 89.2 & 97.0 \\
Total production time (h) & 7.54 & 36.8 \\
Accepted trajectories/h & 291.78 & 59.78 \\
\bottomrule
\end{tabularx}
\end{table}

\paragraph{Production efficiency and human involvement.}
Table~\ref{tab:experiments:production_throughput} shows that BioAgent-World achieves an 89.2\% acceptance rate, compared with 97.0\% for real teleoperation. Processing 2,200 accepted trajectories took only 7.54 hours using eight NVIDIA RTX 4090 GPUs, significantly less than the 36.8 hours needed for real-world data collection. This corresponds to a throughput of 291.78 accepted trajectories per hour, versus 59.78 for real teleoperation, or 4.88 times the real-collection throughput.

Moreover, BioAgent-World requires only  approximately 3 person-hours in total, whereas two real-world collectors remain on duty throughout collection, accounting for 73.6 person-hours. Thus, BioAgent-World improves data-production efficiency and throughput while requiring substantially less human involvement.

\subsection{RQ3: Can Simulation Support BioAgent in Long-Horizon Laboratory Protocols?}
\label{sec:experiments:rq3}

We further evaluate whether BioAgent-World supports long-horizon laboratory protocols beyond isolated skills through a multi-stage procedure integrating navigation, object transport, container manipulation, and liquid handling. BioAgent decomposes the protocol into registered skills, coordinating execution via precondition and postcondition checks while maintaining object identities, spatial relations, and functional states across steps.

As illustrated in Figure~\ref{fig:sim_robot_tasks}, the simulated agent transports reagent tubes across workstations, manipulates a six-well plate, operates a tube-capping device, and performs liquid aspiration and dispensing with a micropipette. This interconnected sequence requires tracking intermediate states and synchronizing manipulation with navigation, demonstrating unified orchestration of interdependent skills in simulation. Moreover, the resulting skill sequence transfers directly to real-robot execution (Figure~\ref{fig:real_robot_tasks}).
\FloatBarrier
\Needspace{4\baselineskip}
\section{Conclusion}
\label{sec:conclusion}

BioAgent-World integrates skill-conditioned digital-twin validation, state-consistent data generation, and active refinement driven by hypothesis disagreements to automate robot training in laboratory environments. Across 11 laboratory skills and three policy backbones, policies trained on our synthetic data retain 87.39--92.22\% of the macro-averaged success rate achieved by real-world data under matched architectures and training budgets. 

Moreover, BioAgent-World only required a total production duration of 7.54 hours to generate 2,200 valid trajectories. In contrast, real-world data collection required a total production time of 36.8 hours.

Despite these gains, the remaining sim-to-real gaps in fine-grained tube-cap manipulation and final plate-lid replacement point to the need for more targeted contact refinement. Future work will also decouple scene construction from active refinement to systematically isolate their individual contributions to transfer performance.
\clearpage

{\small
\bibliographystyle{ieeenat_fullname}
\bibliography{main}
}

\clearpage
\appendix
\setcounter{table}{0}
\setcounter{figure}{0}
\setcounter{equation}{0}
\renewcommand{\thetable}{A\arabic{table}}
\renewcommand{\thefigure}{A\arabic{figure}}
\renewcommand{\theequation}{A\arabic{equation}}

\section{Supplementary Experiments}
\label{app:experiments}

We evaluate two components of BioAgent-World. The first experiment examines
how skill-conditioned trajectory validation affects the validity of accepted
demonstrations (Secs.~3.2 and 3.4 of the main paper). The second compares
active and random observation selection for local scene refinement under a
fixed measurement budget (Sec.~3.3), measuring parameter recovery and checking
the resulting placement targets against independently evaluated simulated
executions.

\subsection{Effect of Skill-Conditioned Trajectory Validation}
\label{app:skill_validation}

\paragraph{Evaluation setup.}
We compare basic trajectory checks with the same checks augmented by
skill-conditioned validation, applying both rules to the same 90 recorded
simulation trajectories. The additional checks evaluate the skill conditions
compiled in Sec.~3.2 and used for trajectory acceptance in Sec.~3.4 of the
main paper. They assess whether execution establishes the required object
states and relations, including those needed by subsequent skills. We report the
number of accepted trajectories, the valid and invalid counts within each
accepted set, and accepted-set validity, defined as the fraction of accepted
demonstrations found valid by independent review.

\begin{table}[!htbp]
\centering
\footnotesize
\setlength{\tabcolsep}{3pt}
\caption{Trajectory-validation ablation on 90 simulated trajectories.
Both rules screen the same candidates. Acceptance rate is the fraction of
candidates accepted; accepted-set validity is the fraction of accepted
trajectories found valid by independent review.}
\label{tab:skill_validation}
\begin{tabular}{@{}lrr@{}}
\toprule
Measure & \shortstack{Basic\\checks} & \shortstack{Basic +\\skill checks} \\
\midrule
Candidate trajectories & 90 & 90 \\
Accepted trajectories & 88 & 78 \\
Acceptance rate (\%) & 97.8 & 86.7 \\
Valid accepted trajectories & 78 & 78 \\
Invalid accepted trajectories & 10 & 0 \\
Accepted-set validity (\%) & 88.6 & 100.0 \\
\bottomrule
\end{tabular}
\end{table}

\paragraph{Results.}
Basic checks accept 88 trajectories: 78 valid demonstrations and 10 invalid
ones (Table~\ref{tab:skill_validation}). Adding skill-conditioned validation
excludes the 10 invalid demonstrations while retaining all 78 valid ones,
raising accepted-set validity from 88.6\% to 100.0\%. The acceptance rate
drops because invalid demonstrations are removed, while the valid count
remains unchanged. This result shows improved demonstration filtering on
the evaluated recordings.

\paragraph{Why object states matter.}
A complete motion record does not necessarily indicate that a skill
succeeded. For example, a lid-pickup trajectory may be recorded in full even
if the lid slips during lifting, leaving the grasped-lid state required for
subsequent placement unsatisfied. By checking these outcome states and
relations, skill-conditioned validation prevents such a trajectory from
being accepted as a valid demonstration solely because it passes the basic
checks.

\subsection{Active Query Selection under a Fixed Measurement Budget}
\label{app:active_query}

We evaluate whether the active query strategy in Sec.~3.3 improves
the measurement efficiency of scene refinement. Using a simulated
tube-placement task with uncertainty in the water bath's position
along the world $X$ axis, we compare active and random query selection
under matched measurement budgets, assessing registration accuracy
and whether the selected placement targets satisfy the skill contract.
We further examine the effect of a local contact-model correction on
placement stability through native-physics execution and trace how
successive queries update the candidate-model weights.

\subsubsection{Experimental setup}
\label{app:active_setup}

\paragraph{Scene hypotheses and execution.}
We specialize the candidate-model representation in Sec.~3.3 to uncertainty
in bath registration. Each model
$M_i=(z_i,\theta_{\mathrm{scene},i},\eta_{\mathrm{robot},i})$ represents a
structural hypothesis, scene parameters, and robot calibration. In this
controlled ablation, the candidates differ only in bath translation,
$(x_0,x_1,x_2)=(0,-8,+8)$\,mm, with a uniform prior $p_i^{(0)}=1/3$.
Each translation moves the bath and its fixed anchors together. The hidden
physical scene uses one of these translations, ensuring that the candidate
set contains the true registration. All other model attributes, robot
calibration, and initial robot and base states are held fixed. A frozen
Monte02 joint-drive diagnostic controller executes the placement skill in
Isaac Sim 6.0.0.1, with the GPU PGS solver and a 2\,ms simulation step on an
RTX 4090. For this discrete registration problem, the
maximum-a-posteriori (MAP) estimate specifies the local bath correction
written to a derived USD layer and the corresponding placement target.

\paragraph{Measurements and comparison protocol.}
We instantiate the local measurement queries of Sec.~3.3 as simulated range
observations. The query set comprises 147 rays arranged in three fixed
$7\times7$ grids over the right opening, left opening, and housing.
Each query returns a scalar range to a USD collision surface, with additive
Gaussian noise of standard deviation $\sigma=1$\,mm. Each ray measurement
counts as one query and costs one unit. We compare active selection with
uniform random selection among the remaining unqueried rays at budgets of
one and three measurements. Both strategies share the candidate set, prior,
observation model, Bayesian update, MAP rule, controller, and physical
evaluation.

\subsubsection{Skill-usability risk and active refinement}
\label{app:active_implementation}

The diagnostic tests evaluate the same skill at three fixed placement targets,
$(g_0,g_1,g_2)=(0,-8,+8)$\,mm. The opening and tube radii are 16 and 14\,mm,
respectively, so the radial-clearance margin in Eq.~(2) of the main paper
becomes
\begin{equation}
\gamma_{ik}=2\,\mathrm{mm}-|x_i-g_k|,
\qquad i,k\in\{0,1,2\}.
\label{eq:app_clearance}
\end{equation}
Geometric screening checks whether $\gamma_{ik}\geq0$ and precedes short
native-physics rollouts with the same diagnostic controller, as described in
Sec.~3.3. Let $Y_{ik}=Y_{s_k}(M_i)\in\{0,1\}$ indicate whether target $k$
satisfies the compiled skill contract under model $M_i$. Contract evaluation
includes the relevant geometric and temporal conditions, such as maintaining
support after gripper release. For the tested configurations, the usability
labels form the $3\times3$ identity matrix, consistent with the execution
results in Table~\ref{tab:app_physical_matrix}.

Let $p_i$ be the current probability of model $M_i$, and let
$a_k(p)=\sum_{i=0}^{2}p_iY_{ik}$ be the predicted probability that target
$k$ satisfies its skill contract. With uniform task weights $w_k=1/3$
and unit symmetric binary loss, the decision risk in Eq.~(4) of the main
paper specializes to
\begin{equation}
R(p)=\frac{1}{3}\sum_{k=0}^{2}\min\{a_k(p),1-a_k(p)\}.
\label{eq:app_risk}
\end{equation}

Let $\mu_i(q)$ denote the range predicted by model $M_i$ for query $q$.
For an observed range $y$, the Gaussian likelihood gives the posterior
\begin{align}
y\mid q,M_i &\sim \mathcal{N}\!\left(\mu_i(q),\sigma^2\right),
\label{eq:app_likelihood}\\
p_i^{q,y}
&=\frac{p_i\exp\!\left[-(y-\mu_i(q))^2/(2\sigma^2)\right]}
{\sum_{j=0}^{2}p_j\exp\!\left[-(y-\mu_j(q))^2/(2\sigma^2)\right]}.
\label{eq:app_posterior}
\end{align}
Following Eq.~(5) of the main paper, active selection maximizes the expected
reduction in decision risk per unit query cost:
\begin{equation}
q^{*}=\underset{q\in\mathcal{Q}_{\mathrm{avail}}}{\arg\max}\;
\frac{R(p)-\mathbb{E}_{y\sim p(y\mid q)}
\!\left[R\!\left(p^{q,y}\right)\right]}{c(q)},
\qquad c(q)=1.
\label{eq:app_query_value}
\end{equation}
Here, $p(y\mid q)$ is the predictive observation distribution and
$p^{q,y}$ is the posterior after observing $y$. Admissible queries satisfy
the budget and execution constraints specified in Sec.~3.3; in this
experiment, previously measured rays are excluded. When full risk evaluation
is costly, the candidate-pair approximation prioritizes queries that
distinguish models with conflicting skill outcomes. This directs measurements
toward scene properties that affect the placement decision.

After acquiring an observation, we update model weights using
Eq.~\eqref{eq:app_posterior}, which instantiates Eq.~(6) of the main paper
for the Gaussian range likelihood. The resulting evidence determines the
local registration correction, whose placement outcome is checked against
the compiled skill conditions.

\subsubsection{Evaluation protocol and results}
\label{app:active_results}

We evaluate each strategy on three hidden translations and three query seeds
(0, 1, and 2), yielding nine decision configurations per strategy and 18 query
sequences in total. The one-query result is a checkpoint of the corresponding
three-query sequence. We measure registration mean absolute error (MAE) from
the selected $X$ translation, exact candidate recovery, and the fraction of
selected truth--target configurations that pass the independent physical
checks described below. For MAE and passing fraction, we first average over
the three truths within each seed, then compute the mean and sample standard
deviation across seeds.

\begin{table}[!htbp]
\centering
\footnotesize
\setlength{\tabcolsep}{2.4pt}
\renewcommand{\arraystretch}{1.08}
\caption{Active and random query selection under matched measurement budgets.
MAE and passing fractions are reported as mean $\pm$ sample standard deviation
across three query seeds, after averaging over three truths within each seed.
Physical outcomes are obtained from the separately executed truth--target
matrix (Table~\ref{tab:app_physical_matrix}); each row contains nine decision
configurations, which reuse these physical outcomes.}
\label{tab:active_query}
\begin{tabular}{@{}lcccc@{}}
\toprule
Strategy & Queries & \shortstack{$X$ MAE\\(mm)} &
\shortstack{Correct\\model} & \shortstack{Audit-passing\\(\%)} \\
\midrule
Random & 1 & $4.44\pm1.54$ & 4/9 & $44.4\pm19.2$ \\
Active & 1 & $1.78\pm1.54$ & 7/9 & $77.8\pm19.2$ \\
\midrule
Random & 3 & $3.56\pm3.08$ & 5/9 & $55.6\pm38.5$ \\
Active & 3 & $0.00\pm0.00$ & 9/9 & $100.0\pm0.0$ \\
\bottomrule
\end{tabular}
\end{table}

After one measurement, active selection recovers the correct candidate in
seven of nine configurations, compared with four for random selection
(Table~\ref{tab:active_query}). This increases the passing fraction by
33.3 percentage points and reduces registration MAE by approximately 60\%.
With three measurements, active selection recovers the correct candidate in
all nine configurations, compared with five for random selection, a difference
of 44.4 percentage points. One active measurement also identifies the correct
candidate more often than three random measurements (7/9 versus 5/9), showing
greater measurement efficiency in this controlled setting.

\subsubsection{Physical execution and validation}
\label{app:active_physics}

\paragraph{Independent success criteria.}
The evaluator checks the placement contract using the final 1\,s of recorded
PhysX states and contacts, requiring at least 45 frames. Success requires nonnegative radial clearance
in the true scene, support-height error below 2\,mm, and maximum drift below
0.6\,mm at both the tube bottom and a second point 117\,mm above it.
The gripper must be open with no finger-pad contact in at least 90\% of
frames, and the holder must support the tube in at least 90\% of frames.
The evaluator and its thresholds remain fixed across methods and contact
settings. This validation checks both geometric feasibility and the
persistent support and release conditions required by the skill.

\paragraph{Contact-model correction.}
The local refinement and revalidation procedure also addresses discrepancies
in contact behavior. Initial tests with the correct target succeeded for two
of the three bath translations. In the $+8$\,mm case, the tube remained
supported after release but oscillated enough to fail the stability check.
We reduced the contact offsets of the tube bottom and support surface from
0.30 to 0.03\,mm as a single-factor intervention, keeping geometry, mass and
inertia, friction, rest offsets, solver settings, time step, and joint
control fixed. These corrected settings were frozen before evaluating the
full execution matrix and applied identically to both query strategies.

Simulator readbacks for this diagnostic case showed the mean support force remained
approximately 0.343\,N, and the corrected case also passed an additional
5\,s observation. The correction increased peak transient penetration at
first impact from approximately 0.015 to 0.214\,mm and peak support force
from 3.02 to 4.90\,N, while the final support, clearance, and stability checks
still passed. 

The physical outcome study thus comprises nine distinct truth--target
trajectories. Three additional diagnostic runs and one extended-observation
control bring the total to 13 native simulation runs. The 18 query sequences
yield 36 checkpoint records across the two budgets, all scored using the
same nine-cell outcome cache. 

\paragraph{Truth--target execution matrix.}
We execute all nine combinations of physical truth and placement target
using the same diagnostic controller (Table~\ref{tab:app_physical_matrix}).
Placement succeeds for all three correct targets and fails for all six
incorrect targets, which trigger the existing slip-stop condition.
The candidate-recovery and audit-passing counts in
Table~\ref{tab:active_query} therefore coincide, although candidate identity
and physical success are assessed independently. Using this same matrix,
a controller that always selects the nominal target succeeds for one of the
three truths, while one given the true target succeeds for all three.

\begin{table}[!htbp]
\centering
\footnotesize
\setlength{\tabcolsep}{3pt}
\caption{Native-physics execution results for the nine truth--target
combinations after applying the common contact correction. Each cell records
one execution with the fixed diagnostic controller, assessed using the
compiled skill conditions. Query seeds reuse these outcomes to score their
selected targets.}
\label{tab:app_physical_matrix}
\begin{tabular}{@{}lccc@{}}
\toprule
\shortstack[l]{Bath translation\\(mm)} &
\shortstack{Target\\0\,mm} &
\shortstack{Target\\$-8$\,mm} &
\shortstack{Target\\$+8$\,mm} \\
\midrule
0\,mm & Success & Failure & Failure \\
$-8$\,mm & Failure & Success & Failure \\
$+8$\,mm & Failure & Failure & Success \\
\bottomrule
\end{tabular}
\end{table}

\subsubsection{Query-by-query refinement}
\label{app:active_trace}

\begin{figure}[!htbp]
\centering
\includegraphics[width=0.95\columnwidth]{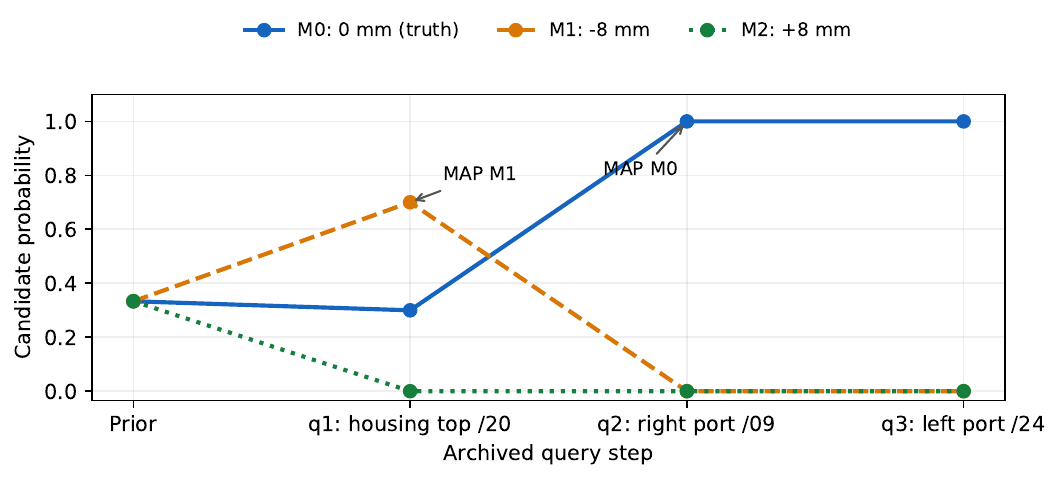}
\caption{Evolution of candidate-model weights in the archived simulated-range
sequence. The true bath translation is $M_0$ (0\,mm). From a uniform prior,
the first noisy observation favors $M_1$ ($-8$\,mm); the second recovers
$M_0$, and the third completes the fixed measurement budget. Weights are
ordered as $(M_0,M_1,M_2)$. Very small positive values appear as zero at the
figure scale.}
\label{fig:section33_weights}
\end{figure}

Figure~\ref{fig:section33_weights} illustrates a representative active-query
sequence reconstructed from the experiment records, with truth $M_0$ and
query seed 1. The first measurement favors the incorrect
candidate $M_1$, but the second shifts the MAP estimate to the true candidate
$M_0$. Table~\ref{tab:section33_query_replay} details the corresponding updates.
Query identities, expected and realized risks, and final weights were recorded
in the trace.
For this replay, ranges and intermediate weights were reconstructed
deterministically from the archived noiseless range bank using the original
observation-noise generator.

\begin{table}[!htbp]
\centering
\footnotesize
\setlength{\tabcolsep}{3pt}
\caption{Archived query sequence for truth $M_0$ and seed 1, with a cost of
one unit per query. Expected risk is the recorded estimate of posterior
risk before the selected measurement is observed; realized risk
is computed after that observation. Posterior weights are ordered as
$(M_0,M_1,M_2)$, and values below $10^{-12}$ are shown as $\simeq0$.}
\label{tab:section33_query_replay}
% Panel 1: measured query and the MAP decision.
\begin{tabular}{@{}clrc@{}}
\toprule
Step & Query & Range (mm) & MAP \\
\midrule
1 & Housing top / 20   & 300.3933 & $M_1$ \\
2 & Right opening / 09 & 300.6518 & $M_0$ \\
3 & Left opening / 24  & 488.9260 & $M_0$ \\
\bottomrule
\end{tabular}
\par\vspace{0.5em}
% Panel 2: posterior state and risk values from the same steps.
\scriptsize
\setlength{\tabcolsep}{2.5pt}
\begin{tabular}{@{}clrr@{}}
\toprule
Step & Posterior $(M_0,M_1,M_2)$ &
\shortstack{Expected\\risk} & \shortstack{Realized\\risk} \\
\midrule
1 & $(0.2998,0.7002,\simeq0)$ & 0.082421 & 0.199841 \\
2 & $(1.0000,\simeq0,\simeq0)$ & $\simeq0$ & $\simeq0$ \\
3 & $(1.0000,\simeq0,\simeq0)$ & $\simeq0$ & $\simeq0$ \\
\bottomrule
\end{tabular}
\end{table}

Starting from an initial risk of $1/3$, the first query yields a posterior
of approximately $(0.299762,0.700238,0)$ and a realized risk of 0.199841,
above the expected posterior risk of 0.082421. The second query resolves the ambiguity in favor of $M_0$.

\end{document}